\def\year{2020}\relax
\documentclass[letterpaper]{article} 
\usepackage{aaai20}  
\usepackage{times}  
\usepackage{helvet} 
\usepackage{courier}  
\usepackage[hyphens]{url}  
\usepackage{graphicx} 
\usepackage{graphicx}  
\usepackage{tabularx}
\usepackage{multirow}
\usepackage{colortbl}
\usepackage{graphicx}
\usepackage{adjustbox}
\usepackage{hhline}
\usepackage{booktabs}
\usepackage{amsmath}

\definecolor{mygray}{gray}{.9}
\DeclareTextFontCommand{\texttth}{\ttfamily\hyphenchar\font=45\relax}
\title{End-to-End Trainable Non-Collaborative Dialog System}
\author{Yu Li, Kun Qian, Weiyan Shi, Zhou Yu\\
University of California, Davis\\ 
\{yooli, kunqian, wyshi, joyu\}@ucdavis.edu 
}
\begin{document}
\maketitle
\begin{abstract}
End-to-end task-oriented dialog models have achieved promising performance on collaborative tasks where users willingly coordinate with the system to complete a given task. While in non-collaborative settings, for example,  negotiation and persuasion, users and systems do not share a common goal. As a result, compared to collaborate tasks, people use social content to build rapport and trust in these non-collaborative settings in order to advance their goals. To handle social content, we introduce a hierarchical intent annotation scheme, which can be generalized to different non-collaborative dialog tasks. Building upon TransferTransfo \cite{wolf2019transfertransfo}, we propose an end-to-end neural network model to generate diverse coherent responses. Our model utilizes intent and semantic slots as the intermediate sentence representation to guide the generation process. In addition, we design a filter to select appropriate responses based on whether these intermediate representations fit the designed task and conversation constraints. 
Our non-collaborative dialog model guides users to complete the task while simultaneously keeps them engaged. 
We test our approach on our newly proposed~\textsc{AntiScam} dataset and an existing~\textsc{PersuasionForGood} dataset. Both automatic and human evaluations suggest that our model outperforms multiple baselines in these two non-collaborative tasks.
\end{abstract}

\section{Introduction}
Considerable progress has been made building end-to-end dialog systems for collaborative tasks in which users cooperate with the system to achieve a common goal. Examples of collaborative tasks include making restaurant reservations and retrieving bus time-table information.
Since users typically have clear and explicit intentions in collaborative tasks, existing systems commonly classify user utterances into pre-defined intents. In contrast, non-collaborative tasks are those where the users and the system do not strive to achieve the same goal. Examples of such tasks include deceiving attackers, persuading users to donate to a cause~\cite{wang2019persuasion}, and negotiating a product price~\cite{he2018decoupling,cao2018emergent}. In these tasks, users often perform complex actions that are beyond a simple set of pre-defined intents. In order to reach a common state, the user and the system need to build rapport and trust which naturally involves off-task content.  Previous work did not model off-task content ~\cite{he2018decoupling}, which may have led to less optimal results. For example, in the persuasion task~\cite{wang2019persuasion}, users would ask the system ``How do you feel about war?" An example of an on-task system response that the system could have made is ``Do you want to make a donation?", which sticks to the task but neglects users' question. However, a better response to such an off-task question is ``War is destructive and pitiless, but you can donate to help child victims of war." This response is better, as it has been found that users are more likely to end the conversation if the system neglects their questions~\cite{yu2017learning}. Therefore, we need to design a system that handles both on-task and off-task information appropriately and in a way that leads back to the system's goal.

To tackle the issue of incoherent system responses to off-task content, previous studies have built hybrid systems to interleave off-task and on-task content. \citeauthor{yu2017learning} used a rule-based dialog manager for on-task content and a neural model for off-task content, and trained a reinforcement learning model to select between these two models based on the dialog context. However, such a method is difficult to train and struggles to generalize beyond the movie promotion task they considered. To tackle these problems, we propose a hierarchical intent annotation scheme that separates on-task and off-task information in order to provide detailed supervision. For on-task information, we directly use task-related intents for representation. Off-task information, on the other hand, is too general to categorize into specific intents, so we choose dialog acts that convey syntax information. These acts, such as ``open question" are general to all tasks.

Previous studies use template-based methods to maintain sentence coherence. However, rigid templates lead to limited diversity, causing the user losing engagement. On the other hand, language generation models can generate diverse responses but are bad at being coherent. We propose Multiple Intents and Semantic Slots Annotation Neural Network (MISSA) to combine the advantages of both template and generation models and takes advantage from the hierarchical annotation at the same time. MISSA follows the TransferTransfo framework~\cite{wolf2019transfertransfo} with three modifications: 
(i) We first concurrently predict user's, system's intents and semantic slots; 
(ii) We then perform conditional generation to improve generated response's coherence. Specifically, we generate responses conditioned on the above intermediate representation (intents and slots);
(iii) Finally, we generate multiple responses with the \textit{nucleus} sampling strategy~\cite{holtzman2019curious} and then apply a response filter, which contains a set of pre-defined constraints to select coherent responses. The constraints in the filter can be defined according to specific task requirements or general conversational rules.

To enrich publicly available non-collaborative task datasets, we collect a new dataset \textsc{AntiScam}, where users defend themselves against attackers trying to collect personal information. As non-collaborative tasks are still relatively new to the study of dialog systems, there are insufficiently many meaningful datasets for evaluation and we hope this provides a valuable example. We evaluate MISSA on the newly collected~\textsc{AntiScam} dataset and an existing~\textsc{PersuasionForGood} dataset. Both automatic and human evaluations suggest that MISSA outperforms multiple competitive baselines.

In summary, our contributions include: (i) We design a hierarchical intent annotation scheme and a semantic slot annotation scheme to annotate the non-collaborative dialog dataset, we also propose a carefully-designed~\textsc{AntiScam} dataset to facilitate the research of non-collaborative dialog systems. (ii) We propose a model that can be applied to all non-collaborative tasks, outperforming other baselines on two different non-collaborative tasks. (iii) We develop an anti-scam dialog system to occupy attacker's attention and elicit their private information for social good. Furthermore, we also build a persuasion dialog system to persuade people to donate to charities. We release the code and data.\footnote{https://gitlab.com/ucdavisnlp/antiscam}

\section{Related Work}
The interest in non-collaborative tasks has been increasing and there have already been several related datasets. 
For instance,~\citeauthor{wang2019persuasion}~\shortcite{wang2019persuasion} collected conversations where one participant persuades another to donate to a charity.~\citeauthor{he2018decoupling}~\shortcite{he2018decoupling} collected negotiation dialogs where buyers and sellers bargain for items for sale on Craigslist. There are many other non-collaborative tasks, such as the turn-taking game~\cite{devault2015toward}, the multi-party game~\cite{asher2016discourse} and item splitting negotiation~\cite{Potts12goal-drivenanswers}. Similar to the~\textsc{AntiScam} dataset proposed in this paper, these datasets contain off-task content and can be used to train non-collaborative dialog systems. 
However, since they are not specifically collected and designed for  non-collaborative tasks, it might be difficult to disentangle the on-task and off-task contents and measure the performance. Therefore, we propose the \textsc{AntiScam} dataset, which is designed to interleave the on-task and off-task contents in the conversation, and can serve as a  benchmark dataset for similar non-collaborative tasks.

To better understand user utterances and separate on-task and off-task content within a conversation, previous work has designed hierarchical annotation schemes for specific domains.~\citeauthor{hardy2002multi}~\shortcite{hardy2002multi} followed the DAMSL scheme\cite{allen1997draft} and annotated a multilingual human-computer dialog corpus with a hierarchical dialog act annotation scheme.~\citeauthor{gupta2018semantic}~\shortcite{gupta2018semantic} used a hierarchical annotation scheme for semantic parsing. Inspired by these studies, our idea is to annotate the intent and semantic slot separately in non-collaborative tasks. We propose a hierarchical intent annotation scheme that can be adopted by all non-collaborative tasks. With this annotation scheme, MISSA is able to quickly build an end-to-end trainable dialog system for any non-collaborative task.

Traditional task-oriented dialog systems~\cite{Young13pomdp-basedstatistical} are usually composed of multiple independent modules, for example, natural language understanding, dialog state tracking~\cite{williams2016dialog,mrkvsic2016neural}, dialog policy manager~\cite{levin2000stochastic}, and natural language generation~\cite{lei2018sequicity}. Conversational intent is adopted to capture the meaning of task content in these dialog systems~\cite{he2018decoupling,zhao2017learning}. In comparison to this work, we use a hierarchical intent scheme that includes off-task and on-task intents to capture utterance meaning. We also train the model in a multi-task fashion to predict decoupled intents and semantic slots. The major defect of a separately trained pipeline is the laborious dialog state design and annotation. In order to mitigate this problem, recent work has explored replacing independent modules with end-to-end neural networks ~\cite{wen2016network,williams2017hybrid,liang2019moss}. Our model also follows this end-to-end fashion.

Over the last few years, we have witnessed a huge growth in non-task-oriented dialog systems~\cite{vinyals2015neural,li2016deep}. Social chatbots such as Gunrock~\cite{Gunrock} were able to maintain a conversation for around ten minutes in an open domain. Recent improvements build on top of the transformer and pre-trained language models~\cite{devlin2018bert,yang2019xlnet,radford2019language},  obtained state-of-the-art results on the~\textsc{Persona-Chat} dataset~\cite{wolf2019transfertransfo}. Pre-trained language models are proposed to build task-oriented dialog systems to drive the progress on leveraging large amounts of available unannotated data.~\cite{budzianowski2019hello}. Similarly, our approach is also built on top of the TransferTransfo framework~\cite{wolf2019transfertransfo}.~\citeauthor{budzianowski2019hello}~\shortcite{budzianowski2019hello} focused on collaborative tasks~\cite{budzianowski2018multiwoz}. We target non-collaborative tasks instead.

Another line of work interleaves on-task and off-task content by building a hybrid dialog system that combines a task-oriented model and a non-task-oriented model~\cite{yu2017learning,papaioannou2017hybrid}. In these studies, task-oriented systems and non-task-oriented systems are designed separately and both systems generate candidate responses. A selector is then designed to choose an appropriate output from the candidate responses~\cite{yu2017learning} and a connector to combine two response candidates~\cite{zhao2018sogo,baheti2018generating}. Compared with these works, MISSA is end-to-end trainable and thus easier to train and update.

\begin{table}[htb!]
\centering
\begin{adjustbox}{width=\columnwidth}
\begin{tabular}{|l|c|l|}

    \hline
    \multirow{12}{*}{On-task} & \multirow{3}{*}{\textsc{AntiScam}}& \textit{elicitation}\\ \cline{3-3}
  &  &  \textit{providing\_information} \\ \cline{3-3}
  &  & \textit{refusal} \\ \cline{2-3}
  & & \textit{agree\_donation}\\ \cline{3-3}
  & & \textit{disagree\_donation}\\ \cline{3-3}
  & & \textit{disagree\_donation\_more}\\ \cline{3-3}
  & & \textit{ask\_donation\_amount} \\ \cline{3-3}
  & \textsc{Persuasion-}& \textit{ask\_donate\_more} \\ \cline{3-3}
  & \textsc{-ForGood}& \textit{proposition\_of\_donation} \\ \cline{3-3}
  & & \textit{er\_confirm\_donation} \\ \cline{3-3}
  & & \textit{ee\_confirm\_donation}\\ \cline{3-3}
  & & \textit{provide\_donation\_amount}\\ \cline{1-3}
  \multicolumn{2}{|c|}{\multirow{12}{*}{Off-task}} &  \textit{open\_question}\\ \cline{3-3}         
  \multicolumn{2}{|l|}{} & \textit{yes\_no\_question} \\ \cline{3-3}         
  \multicolumn{2}{|l|}{} & \textit{negative\_answer} \\ \cline{3-3}        
  \multicolumn{2}{|l|}{} & \textit{positive\_answer} \\ \cline{3-3}      
  \multicolumn{2}{|l|}{} & \textit{responsive\_statement} \\ \cline{3-3}      
  \multicolumn{2}{|l|}{} & \textit{nonresponsive\_statement} \\ \cline{3-3}          
  \multicolumn{2}{|l|}{} & \textit{greeting} \\ \cline{3-3}       
  \multicolumn{2}{|l|}{} & \textit{thanking} \\ \cline{3-3}       
  \multicolumn{2}{|l|}{} & \textit{respond\_to\_thank} \\ \cline{3-3}         
  \multicolumn{2}{|l|}{} & \textit{apology} \\ \cline{3-3}      
  \multicolumn{2}{|l|}{} & \textit{closing} \\ \cline{3-3}      
  \multicolumn{2}{|l|}{} & \textit{hold} \\ \cline{1-3}      

\end{tabular}
\end{adjustbox}
\caption{Hierarchical intent annotation scheme on both \textsc{AntiScam} dataset and \textsc{PersuasionForGood} dataset. The On-task intents are task-specific while the Off-task intents are general for different non-collaborative tasks.
  }
\label{Intent annotation scheme}
\end{table}
\section{Non-Collaborative Task Annotation Scheme}
To decouple syntactic and semantic information in utterances and provide detailed supervision, we design a hierarchical intent annotation scheme for non-collaborative tasks. We first separate on-task and off-task intents. As on-task intents are key actions that can vary among different tasks, we need to specifically define on-task intents for each task. On the other hand, since off-task content is too general to design task-specific intents, we choose common dialog acts as the categories. The advantage of this hierarchical annotation scheme is apparent when starting a new non-collaborative task: we only need to focus on designing the on-task categories and semantic slots which are the same as traditional task-oriented dialog systems. Consequently, we don't have to worry about the off-task annotation design since the off-task category is universal.

In the intent annotation scheme shown in Table~\ref{Intent annotation scheme}, we list the designed intent annotation scheme for the newly collected~\textsc{AntiScam} dataset and the~\textsc{PersuasionForGood} dataset. We first define on-task intents for the datasets, which are key actions in the task. Since our~\textsc{AntiScam} focuses on understanding and reacting towards elicitations, we define \textit{elicitation}, \textit{providing\_information} and \textit{refusal} as on-task intents. In the~\textsc{PersuasionForGood} dataset, we define nine on-task intents in Table~\ref{Intent annotation scheme} based on the original~\textsc{PersuasionForGood} dialog act annotation scheme. All these intents are related to donation actions, which are salient on-task intents in the persuasion task. The off-task intents are the same for both tasks, including six general intents and six additional social intents. General intents are more closely related to the syntactic meaning of the sentence (\textit{open\_question}, \textit{yes\_no\_question}, \textit{positive\_answer}, \textit{negative\_answer}, \textit{responsive\_statement}, and \textit{nonresponsive\_statement}) while social intents are common social actions (\textit{greeting}, \textit{closing}, \textit{apology}, \textit{thanking},\textit{respond\_to\_thank}, and \textit{hold}).

For specific tasks, we also design a semantic slot annotation scheme for annotating sentences based on their semantic content. We identify 13 main semantic slots in the anti-scam task, for example, credit card numbers. We present a detailed semantic slot annotation in Table~\ref{Semantic slot annotation scheme}. Following~\citeauthor{wang2019persuasion}, we segment each conversation turn into single sentences and then annotate each sentence rather than turns.

\begin{table}[htb]
\centering
\begin{adjustbox}{width=\columnwidth}
\begin{tabular}{ll}
    \hline
    Annotation      & Examples  \\
    \hline
    \hline
    \textit{order\_detail}    & ``Your order will arrive by Thursday afternoon."  \\
    \textit{order\_update}    & ``Should I cancel your order?"\\
    \textit{payment}         & ``Was payment made with a Visa card?"   \\
    \textit{name}            & ``Can you give me your name?"  \\
    \textit{identity}        & ``I am from Amazon customer service."  \\
    \textit{address}         & ``Can confirm your billing address for me?" \\
    \textit{phone\_num}       & ``What is your phone number?" \\
    \textit{card\_info}       &  ``I need the credit card info please" \\
    \textit{card\_num}        &  ``Can you verify your card number for me?" \\
    \textit{card\_cvs}        & ``Next, I will need the CVS number from that card."   \\
    \textit{card\_date}       &  ``Can I have expiration date?" \\
    \textit{account\_detail}  &  ``This is to confirm your account." \\
    \textit{others}          &  ``How are you doing?" \\
    \hline
\end{tabular}
\end{adjustbox}
\caption{ \textsc{AntiScam}'s semantic slot annotation scheme.
  }
\label{Semantic slot annotation scheme}
\end{table}

\section{Datasets}
We test our approach on two non-collaborative task datasets: the~\textsc{AntiScam} dataset and the~\textsc{PersuasionForGood} dataset~\cite{wang2019persuasion}. Both datasets are collected from the Amazon Mechanical Turk platform in the form of typing conversations and off-task dialog is interleaved in the dialog.

\subsection{~\textsc{AntiScam} Dataset}
To enrich available non-collaborative task datasets, we created a corpus of human-human anti-scam dialogs in order to learn human elicitation strategies. We chose a popular Amazon customer service scam scenario to collect dialogs between users and attackers who aim to collect users information. We posted a role-playing task on the Amazon Mechanical Turk platform and collected a typing conversation dataset named \textsc{AntiScam}. We collected 220 human-human dialogs. The average conversation length is 12.45 turns and the average utterance length is 11.13 words. Only 172 out of 220 users successfully identified their partner as an attacker, suggesting that the attackers are well trained and not too easily identifiable. We recruited two expert annotators who have linguistic training to annotate 3,044 sentences in 100 dialogs, achieving a 0.874 averaged weighted kappa value.

\subsection{~\textsc{PersuasionForGood} Dataset}
The \textsc{PersuasionForGood} dataset~\cite{wang2019persuasion} was collected from typing conversations on Amazon Mechanical Turk platform. Two workers were randomly paired, one was assigned the role of persuader, the other was persuadee. The goal of the persuader was to persuade the persuadee to donate a portion of task earning to a specific charity. The dataset consists of 1,017 dialogs, where 300 dialogs are annotated with dialog acts. The average conversation length is 10.43, the vocabulary size is 8,141. Since the original ~\textsc{PersuasionForGood} dataset is annotated with dialog acts, we select the on-task dialog acts as on-task intents shown in Table~\ref{Intent annotation scheme}, and categorize the other dialog acts into our pre-defined off-task intents.

\begin{figure*}[htb!]
\centering
\includegraphics[width=6.5in,height=3.011in]{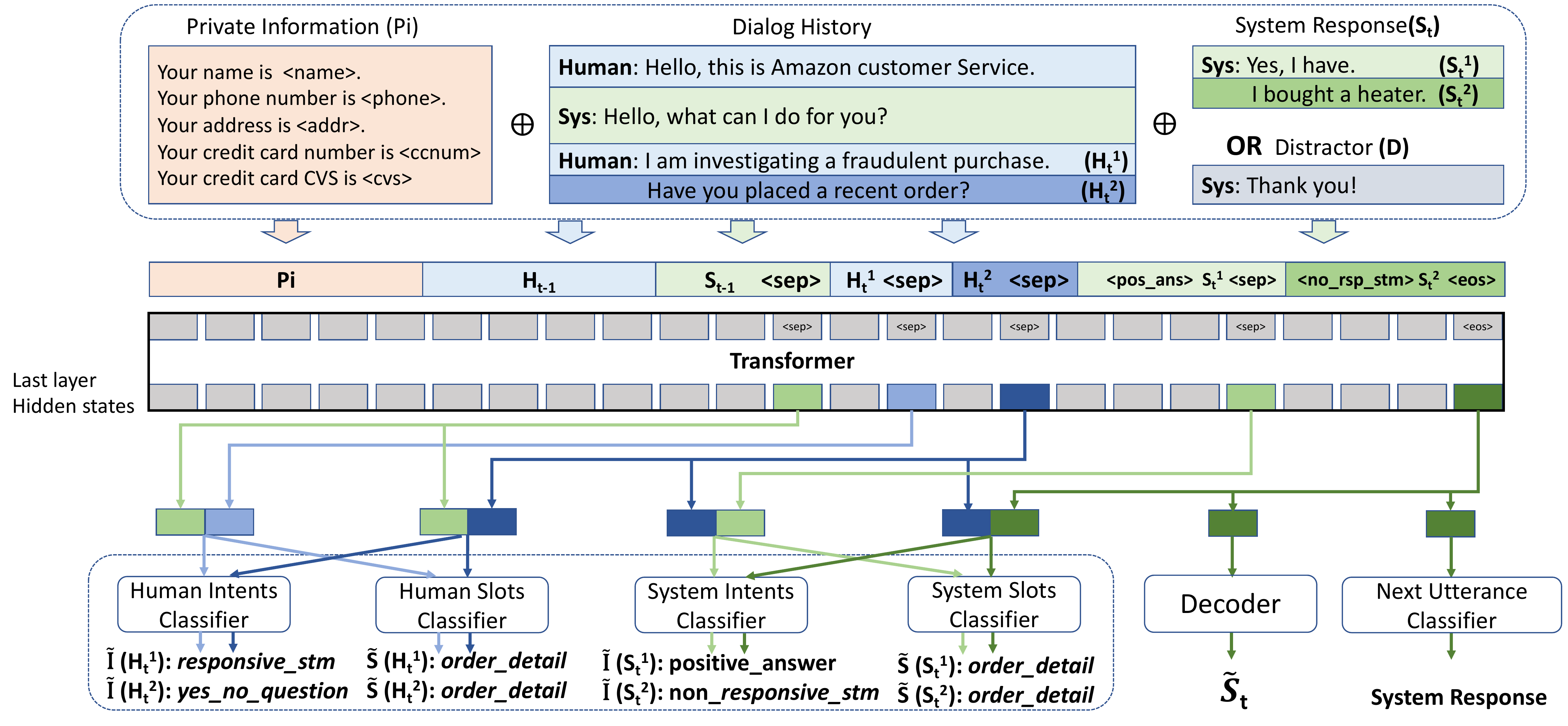}
\caption{The training phase overview of MISSA on~\textsc{AntiScam} dataset, the input consists of three parts: private information, dialog history, and an appended next utterance. We concatenate the last hidden states at $<$sep$>$ tokens with the last hidden states at the end of the last utterance to predict intents and semantic slots for corresponding sentences. We can predict multiple intents and semantic slots for each human utterance and system response. During testing, the appended response and distractor are removed.}
\label{Model}
\end{figure*}

\section{Model}
\subsection{Background}
The TransferTransfo framework was proposed to build open domain dialog systems. ~\citeauthor{wolf2019transfertransfo}~\shortcite{wolf2019transfertransfo} fine-tuned the generative pre-training model (GPT)~\cite{radford2018improving} with the \textsc{PERSONA-CHAT} dataset~\cite{zhang2018personalizing} in a multi-task fashion, where the language model objective is combined with a next-utterance classification task. The language model's objective is to maximize the following likelihood for a given sequence of tokens, $X = \{x_1,\dots,x_n\}$:
\begin{equation}
    {\mathcal{L}_{LM}}(X) = \sum^n_{i=1}{\log P(x_i|x_0,\dots,x_{i-1})}
    \label{eq1}
\end{equation}

The authors also trained a classifier to distinguish the correct next-utterance appended to the input human utterances from a set of randomly selected utterance distractors. In addition, they introduced dialog state embeddings to indicate speaker role in the model. The model significantly outperformed previous baselines over both automatic evaluations and human evaluations in social conversations. Since the TransferTransfo framework performs well in open domain, we adapt it for non-collaborative settings. We keep all the embeddings in the framework and train the language model and next-utterance classification task in a multi-task fashion following TransferTransfo.  

We make two major changes: (1) To address the problem that TransferTransfo is originally designed for an open domain without explicit intents and regulations, we add two intent classifiers and two semantic slot classifiers to classify the intents and semantic slots for both human utterances and system responses as an effort to incorporate the proposed hierarchical intent and semantic slot annotation for non-collaborative tasks. 
(2) In dialog systems, multiple generated responses can be coherent under the current context. Generating diverse responses has proven to be an enduring challenge. To increase response diversity, we sample multiple generated responses and choose an appropriate one according to a set of pre-defined rules.

\subsection{Intent and Semantic Slot Classifiers}
We train MISSA in a multi-task fashion. In addition to the language model task and the next-utterance prediction task, we also use separate classifiers to predict the intents and semantic slots of both human utterances and system responses. The intent classifier and semantic slot classifier for human utterances capture the semantic and syntactic meaning of human utterances, providing information to select the appropriate response among response candidates while the classifiers for the system intents and semantic slots are designed to help select an appropriate next-sentence. We describe response filtering in the corresponding subsection. Classifiers are designed as the following equation:
\begin{equation}
    p(L^i_{t}) = softmax(\left[\begin{array}{l}
         h^l_{t-1}\\
         h^i_{t}\\
         \end{array} \right ] \cdot W_{2h})
    \label{eq2}
\end{equation}
where $L^i_{t}$ is the intent or semantic label of $i$-th sentence at turn $t$. $h^l_{t-1}$ is the hidden states at the end of last sentence in turn $t-1$, $h^i_{t}$ is the last hidden states at the end of $i$-th sentence in turn $t$. $W_{2h}$ are weights learned during training.

MISSA is able to classify multiple intents and multiple semantic slots in a single utterance with these classifiers. Figure~\ref{Model} shows how it works on the \textsc{AntiScam} dataset. Specifically, we set a special token \textit{$<$sep$>$} at the end of each sentence in an utterance (an utterance can consist of multiple sentences). Next, we pass the token's position information to the transformer architecture and obtain the representation of the position (represented as colored position at last layer in Figure~\ref{Model}). After that, we concatenate the embeddings at these position with the hidden states of last sentence. We pass these concatenated representations to the intent classifier and the slot classifier to obtain an intent and a semantic slot for each sentence in the utterance. As shown in Figure~\ref{Model}, the loss function ${\mathcal{L}}$ for the model combines all the task losses:
\begin{equation}
\begin{split}
{\mathcal{L}} = &\lambda_{LM}{\mathcal{L}_{LM}} + \lambda_{I_h}{\mathcal{L}_{I_h}} +\lambda_{S_h}{\mathcal{L}_{S_h}} + \lambda_{I_s}{\mathcal{L}_{I_s}}\\
    &+ \lambda_{S_s}{\mathcal{L}_{S_s}} + \lambda_{nup}{\mathcal{L}_{nup}}
    \label{eq3}
\end{split}
\end{equation}
where ${\mathcal{L}_{LM}}$ is the language model loss, ${\mathcal{L}_{I_h}}$, ${\mathcal{L}_{S_h}}$, ${\mathcal{L}_{I_s}}$, and ${\mathcal{L}_{S_s}}$ are losses of intent and slots classifiers, ${\mathcal{L}_{nup}}$ is next-utterance classification loss. $\lambda_{LM}$, $\lambda_{I_h}$, $\lambda_{S_h}$, $\lambda_{I_s}$, $\lambda_{S_s}$, and $\lambda_{nup}$ are the hyper-parameters that control the relative importance of every loss.

\subsection{Response Generation}
MISSA can generate multiple sentences in a single system turn. Therefore, we perform system generation conditioned on predicted system intents. More specifically, during the training phase, in addition to inserting a special \textit{$<$sep$>$} token at the end of each sentence, we also insert the intent of the system response as special tokens at the head of each sentence in the system response. For example, in Figure~\ref{Model}, we insert a $<$\textit{pos\_ans}$>$ token at the head of $S_t^1$, which is the system response in green. We then use a cross entropy loss function to calculate the loss between the predicted token and the ground truth intent token. During the testing phase, the model first generates a special intent token, then after being conditioned on this intent token, the model keeps generating a sentence until it generates a \textit{$<$sep$>$} token. After that, the model continues to generate another intent token and another sentence until it generates an \textit{$<$eos$>$} token.

\subsection{Response Filtering}
Since we only perform conditional generation, a type of soft constraint on the predicted intent of system response, the system can still generate samples that violate simple conversation regulations, such as eliciting information that has already been provided. These corner cases may lead to fatal results in high-risk tasks, for example, health care and education. To improve the robustness of MISSA and improve its ability to generalize to more tasks, we add a response filtering module after the generation. With the \textit{nucleus} sampling strategy~\cite{holtzman2019curious}, MISSA is able to generate multiple diverse candidate responses with different intents and semantic slots. We then adopt a task-specific response filtering policy to choose the best candidate response as the final output. In our anti-scam scenario, we set up a few simple rules to filter out some unreasonable candidates, for instance, eliciting the repeated information. The filtering module is easily adaptable to different domains or specific requirements, which makes our dialog system more controllable.

\section{Experiments}
We evaluate MISSA on two non-collaborative task datasets. ~\textsc{AntiScam} aims to build a dialog system that occupies the attacker's attention and elicits the attacker's information while ~\textsc{PersuasionForGood}~\cite{wang2019persuasion} aims to build a dialog system that persuades people to donate to a charity. We use $80\%$ data for training, $10\%$ data for validation, and $10\%$ data for testing. More training details are presented in Appendix.

\subsection{Baseline Models}
We compare MISSA mainly with two baseline models:
\begin{itemize}
    \item \textbf{TransferTransfo} The vanilla TransferTransfo framework is compared with MISSA to show the impact and necessity of adding the intent and slot classifiers. We follow the original TransferTransfo design \cite{wolf2019transfertransfo} and train with undelexicalized data.
    \item \textbf{Hybrid} Following~\citeauthor{yu2017learning}~\shortcite{yu2017learning}, we also build a hybrid dialog system by combining vanilla TransferTransfo and MISSA. Specifically, we first determine if the human utterances are on-task or off-task with human intent classifier. If the classifier decides that the utterance is on-task, we choose the response from MISSA; otherwise, we choose the response from vanilla TransferTransfo baseline.
\end{itemize}
In addition, we perform ablation studies on MISSA to show the effects of different components. 
\begin{itemize}
    \item \textbf{MISSA-sel} denotes  MISSA without response filtering.
     \item \textbf{MISSA-con} denotes MISSA leaving out the intent token at the start of the response generation.
\end{itemize}

\begin{table*}[htb!]
\centering
\setlength{\tabcolsep}{1.5mm}{ 
\begin{tabular}{l|ccccc|ccccc}
    \hline
    & \multicolumn{5}{c|}{Automatic Evaluation Metrics} & \multicolumn{5}{c}{Human Evaluation Metrics} \\
    
    Model& PPL & RIP & RSP & ERIP & ERSP & Fluency & Coherence & Engagement& Length & TaskSuc\\
    \hline
    \hline
    TransferTransfo & 32.96 & 34.8\% & 46.0\% & \textbf{48.0\%} & 56.3\%  & 3.48 & 2.85 & 2.68& 8.5 & 1.025\\
    \hline
    Hybrid & - & 32.0\% & 44.0\% & 45.7\% & 55.3\%  & 3.25 & 2.76 & 2.60& 8.2 & 0.975\\
    \hline
    MISSA & \textbf{21.07} & \textbf{35.1\%} & \textbf{46.6\%} & 47.2\% & \textbf{58.6\%}  & \textbf{4.18} &\textbf{3.75}& 3.69& \textbf{14.9} &1.294\\
    \hline
    MISSA-sel & 30.54 & 31.6\% & 42.4\%& 44.2\% & 53.8\%  & 3.60&2.92 &2.87& 9.9&1.000\\
    \hline
    MISSA-con & 24.46 & 33.8\% & 45.6\% & 46.0\% & 57.3\% &3.78&3.68&\textbf{3.78}& 14.8&\textbf{1.341}\\
    \hline
\end{tabular}}
\caption{Experiments results with both automatic and human evaluation on~\textsc{AntiScam} dataset.
  }
\label{table:Experiments}
\end{table*}

\subsection{Automatic Evaluation Metrics}

    \noindent \textbf{Perplexity} Since the canonical measure of a good language model is perplexity, which indicates the error rate of the expected word. We choose perplexity to evaluate the model performance.
    
    \noindent \textbf{Response-Intent Prediction (RIP) $\&$ Response-Slot Prediction (RSP)} Different from open-domain dialog systems, we care about the intents of the system response in non-collaborative tasks as we hope to know if the system response satisfies user intents. For example, in the anti-scam task, if the attacker elicits information from the system, we need to know if the system refuses or agrees to provide the information. Therefore we care about intent prediction for the generated system response. Since our baselines are more suited for social chat as they cannot produce system intents, we use the system intent and slot classifiers trained in our model to predict their responses' intents and slots. The intent predictor achieves a $84\%$ accuracy and the semantic slot predictor achieves $77\%$ on the~\textsc{AntiScam} dataset. Then we compare the predicted values with human-annotated ground truth in the dataset to compute the response-intent prediction (RIP) and response-slot prediction (RSP).

    \noindent \textbf{Extended Response-Intent Prediction (ERIP) $\&$ Extended Response-Slot Prediction (ERSP)} 
    With \textit{Response-Intent Prediction}, we verify the predicted intents to evaluate the coherence of the dialog. However, the real mapping between human-intent and system-intent is much more complicated as there might be multiple acceptable system-intents for the same human-intent. Therefore, we also design a metric to evaluate if the predicted system-intent is in the set of acceptable intents.
    Specifically, we estimate the transition probability $p(I_i|I_j)$ by counting the frequency of all the bi-gram human-intent and system-intent pairs in the training data. During the test stage, if the predicted intent matches the ground truth, we set the score as 1, otherwise we set the score as $p(I_{predict}|I_i)$ where $I_i$ is the intent of the input human utterance.
    We then report the average value of those scores over turns as the final extended response-intent prediction result.


\begin{table*}[htb!]
\centering
\small
\begin{tabularx}{\textwidth}{ll}
\toprule
\hline
\hspace{2mm}(A) Att: Attacker \ \ \ \ \ \ \   System: TransferTransfo & \hspace{2mm}(B) Att: Attacker \ \ \ \ \ \ \   System: MISSA\hspace{70mm}\\
\hline
\begin{tabular}{lp{70mm}}
Att: & Can I have your card number? \\
Sys: & \textbf{5110-xxxx-xxxx-8166} \\
Att: & Got it. I also need the billing address\\
Sys: & \textbf{What is your name? My name is Tom.}\\
Att: & I am Jerry, as I mentioned before. I need your address to proceed the refund.\\
Sys: &\textbf{What is the phone number there?}\\
Att: & Could you give me your phone number?\\
\end{tabular}
&
\begin{tabular}{lp{70mm}} \\
Att: & Can I have your card number? \\
Sys: & \textbf{Why would you need my credit card number?} \\
Att: & I can track your order easily.\\
Sys: & \textbf{Why can't you use my address?}\\
Att: & Well, both of them are needed.\\
Sys: &\textbf{Alright, it is 5110-xxxx-xxxx-8166. May I have your number in case we disconnect?}\\
Att: &xxx-xxx-xxxx\\
\end{tabular}\\

\hline
\hline
\hspace{2mm}(C) Att: Attacker \ \ \ \ \ \ \   System: Hybrid & \hspace{2mm}(D) Att: Attacker \ \ \ \ \ \ \   System: MISSA\hspace{70mm}\\
\hline

\begin{tabular}{lp{70mm}}

Att: & Hi, my name is Jerry.\\
Sys: & \textbf{Hello, who is this?}\\
&...\\
Sys: & \textbf{Why do you need my address?}\\
Att: & I need your address to verify your order.\\
Sys: & \textbf{I don't have my order number with me. Why do you need my order number to verify my order?}\\
\end{tabular}
&
\begin{tabular}{lp{70mm}}
Att: & Hi, my name is Jerry.\\
Sys: & \textbf{Hi, Jerry. what are you calling me about?}\\
&...\\
Sys: & \textbf{Why do you need my address?}\\
Att: & I need your address to verify your order.\\
Sys: & \textbf{You should have it on file already}\\
\end{tabular}\\
\bottomrule
\end{tabularx}
\caption{
Examples of human-system dialogs, where systems are trained on \textsc{AntiScam} dataset. System responses are bolded.
}
\label{table:case study}
\end{table*}

\subsection{Human Evaluation Metrics}
Automatic metrics only validate the system’s performance on a single dimension at a time. The ultimate holistic evaluation should be conducted by having the trained system interact with human users. Therefore we also conduct human evaluations for the dialog system built on ~\textsc{AntiScam}. 
We test our models and baselines with 15 college-student volunteers. Each of them is asked to pretend to be an attacker and interact with all the models for at least three times to avoid randomness. We in total collect 225 number of dialogs.
Each time, volunteers are required to use similar sentences and strategies to interact with all five models and score each model based on the metrics listed below at the end of the current round. Each model receives a total of 45 human ratings, and the average score is reported as the final human-evaluation score. 
In total, we design five different metrics to assess the models' conversational ability whilst interacting with humans. The results are shown in Table~\ref{table:Experiments}.

\noindent\textbf{Fluency} Fluency is used to explore different models' language generation quality.

\noindent\textbf{Coherence} Different from single sentence's fluency, coherence focuses more on the logical consistency between sentences in each turn.

\noindent\textbf{Engagement} In the anti-scam scenario, one of our missions is to keep engaging with the attackers to waste their time. So we directly ask volunteers (attackers) to what extend they would like to continue chatting with the system.

\noindent\textbf{Dialog length (Length)} Engagement is a subjective metric. Anti-scam system's goal is to engage user in the conversation longer in order to limit their harm to other potential victims. So we count the dialog length as another metric to evaluate system performance.

\noindent \textbf{Task Success Score (TaskSuc)} The other goal of the anti-scam system is to elicit attacker's personal information. 
We count the average type of information (name, address and phone number) that the system obtained from attackers as the task success score.

\section{Results and Analysis}
Table~\ref{table:Experiments} presents the main experiment results on~\textsc{AntiScam} dataset, for both automatic evaluation metrics and human evaluation metrics. The experiment results on~\textsc{PersuasionForGood} are shown in Table~\ref{Experiments2}. We observe that MISSA outperforms two baseline models (TransferTransfo and hybrid model) on almost all the metrics on both datasets. For further analysis, examples of real dialogs from the human evaluation are presented in Table~\ref{table:case study}.

Compared to the first TransferTransfo baseline, MISSA outperforms the TransferTransfo baseline on the on-task contents.
From Table ~\ref{table:Experiments}, we observe that MISSA maintains longer conversations (14.9 turns) compared with TransferTransfo  (8.5 turns), which means MISSA is better at maintaining the attacker's engagement. 
MISSA also has a higher task success score (1.294) than TransferTransfo (1.025), which indicates that it elicits information more strategically. 
In the top two dialogs (A and B) that are shown in Table~\ref{table:case study}, both attackers were eliciting a credit card number in their first turns. TransferTransfo directly gave away the information, while MISSA replied with a semantically-related question ``why would you need my credit card number?" 
Furthermore, in the next turn, TransferTransfo ignored the context and asked an irrelevant question ``what is your name?'' while MISSA was able to generate the response ``why can't you use my address?'', which is consistent to the context.
We suspect the improved performance of MISSA comes from our proposed annotation scheme: the semantic slot information enables MISSA to keep track of the current entities, and the intent information helps MISSA to maintain coherency and prolong conversations.

Compared to the hybrid model baseline, MISSA performs better on off-task content. As shown in the bottom two dialogs in Table~\ref{table:case study}, attackers in both dialogs introduced their names in their first utterances. MISSA recognized attacker's name, while the hybrid model did not. We suspect it is because the hybrid model does not have the built-in semantic slot predictor. In the second turn, both attackers were explaining the reason of requesting the billing address previously. With semantic slot information, MISSA can easily understand the attacker; but the hybrid model misunderstands that the attacker was talking about the order number, possibly because the token ``order'' appeared in the attacker's utterance. We suspect that the hybrid model's bad performance on the off-task content leads to its low coherence rating (2.76) and short dialog length (8.2).

To explore the influence of the intent-based conditional response generation method and the designed response filter, we perform an ablation study. The results are shown in Table~\ref{table:Experiments}. We find that MISSA has higher fluency score and coherence score than MISSA-con (4.18 vs 3.78 for fluency, and 3.75 vs 3.68 for coherence), which suggests that conditioning on the system intent to generate responses improves the quality of the generated sentences.  Compared with MISSA-sel, MISSA achieves better performance on all the metrics. For example, the engagement score for MISSA is 3.69 while MISSA-sel only has 2.87. This is because the response filter removed all the incoherent responses, which makes the attacker more willing to keep chatting. The ablation study shows both the conditional language generation mechanism and the response filter are essential to MISSA's good performance.

We also apply our method to  the~\textsc{PersuasionForGood} dataset. As  shown in Table~\ref{Experiments2}, MISSA and its variants outperform the TransferTransfo and the hybrid models on all evaluation metrics. Such good performance indicates MISSA can be easily applied to a different non-collaborative task and achieve good performance. Particularly, MISSA achieves the lowest perplexity, which confirms that using conditional response generation leads to  high quality responses. Compared with the result on~\textsc{AntiScam} dataset, MISSA-con performs the best in terms of RIP and ERIP. We suspect the underlying reason is that there are more possible responses with the same intent in~\textsc{PersuasionForGood} than in~\textsc{AntiScam}. This also suggests that we should adjust the model structure according to the nature of the dataset.

\begin{table}[htb!]
\centering
\begin{tabular}{l|ccc}
    \hline
     Model& PPL & RIP & ERIP \\
    \hline
    \hline
    TransferTransfo & 26.37 & 41.9\% & 48.0\% \\
    \hline
    Hybrid & - & 37.3\% & 44.1\% \\
    \hline
    MISSA & \textbf{19.91} & 43.1\%  & 51.6\% \\
    \hline
    MISSA-sel & 24.00 & 42.0\% & 48.9\% \\
    \hline
    MISSA-con & 20.43 & \textbf{43.8}\% & \textbf{52.6}\%\\
    \hline
\end{tabular}
\caption{Automatic evaluation results  on~\textsc{PersuasionForGood} dataset.
  }
\label{Experiments2}
\end{table}

\begin{table*}[htb!]
\centering
\small
\begin{tabular}{p{2\columnwidth}}
    \hline
    \textbf{Instructions for the attacker}\\
    \hline
    Your name is Norman Paul.\\
    Your phone number is 800-xxx-8892\\
    Your location is 2100 Kings ct, x City, x State, xxxxx (Zipcode)\\

    You are a hacker and you want to get others' information as much as possible.\\
    You are pretending to be an Amazon Customer Service staff.\\
    You want him/her to tell you his/her card number, CVS number on the card, expire date, name and billing address etc.\\

    The information of the other person you have already known:\\
    \quad His name is Jim Lee.\\
    \quad He bought something on Amazon.com.\\
    You may get a bonus if you get his credit card information successfully.\\
    \hline
    \hline
    \textbf{Instructions for the user}\\
    \hline
    Your name is Jim Lee.\\
    Your credit card number is 5110-xxxx-xxxx-8166\\
    Your CVS number of credit card is 380\\
    Your credit card expire date is 05/25\\
    Your phone number is 350-xxx-2988\\
    Your billing address is xxx El Ave, Apt 311, City, State, Zipcode\\

    You bought a heater on Amazon.com and it has not arrived.\\

    Now you get a phone call from an unknown number\\
    Stage1:\\
    \quad You respond to the other person ordinarily.\\
    \quad After several turns, you need to judge if the other person is an attacker. If you think he/she is an attacker, go to stage2. Otherwise, you stay at stage1.\\

    Stage2:\\
    \quad You are not supposed to give your information to him/her.\\
    \quad You are trying to get his/her information as mush as possible, including name, phone number, location, etc.\\
    \quad If you can not get his/her information, try to talk with him/her and waste his/her time as long as possible.\\
    \quad You may get a bonus if you can get his information correctly.\\
    \hline
\end{tabular}
\caption{Instructions for attackers and users on Amazon Mechanical Turk.
  }
\label{Instructions for workers}
\end{table*}

\section{Conclusion and Future Work}
We propose a general dialog system pipeline to build non-collaborative dialog systems, including a hierarchical annotation scheme and an end-to-end neural response generation model called MISSA.  
With the hierarchical annotation scheme, we can distinguish on-task and off-task intents. MISSA takes both on and off-task intents as supervision in its training and thus can deal with diverse user utterances in non-collaborative settings. Moreover, to validate MISSA's performance, we create a non-collaborate dialog dataset that focuses on deterring phone scammers. 
MISSA outperforms all baseline methods in terms of fluency, coherency, and user engagement on both the newly proposed anti-scam task and an existing persuasion task. However, MISSA still produces responses that are not consistent with their distant conversation history as GPT can only track a limited history span. In future work, we plan to address this issue by developing methods that can effectively track longer dialog context.

\section{Acknowledgements}
This work was supported by DARPA ASED Program HR001117S0050. The U.S. Government is authorized to reproduce and distribute reprints for governmental purposes not withstanding any copyright notation therein. The views and conclusions contained herein are those of the authors and should not be interpreted as necessarily representing the official policies, either expressed or implied, of DARPA or the U.S. Government.

\section{Appendix}

\label{sec:appendix}

\begin{table*}[htb!]
\centering
\small
\begin{tabular}{lp{8.95cm}ll}
    \hline
    Role & Sentences & Intent & Semantic Slot\\
    \rowcolor{mygray}
    Attacker & Hi. & \textit{greeting} & \textit{others}\\
    \rowcolor{mygray}
    Attacker & I'm $<$name$>$ with Amazon's Distribution Center. & \textit{greeting}& \textit{name}\\
    User & Hello $<$name$>$ I'm $<$name$>$ how are you today? & \textit{greeting} & \textit{name}\\
    \rowcolor{mygray}
    Attacker & I'm doing very well, thank you for asking. & \textit{greeting} & \textit{others}\\
    \rowcolor{mygray}
    Attacker & How did you enjoy your recent Amazon purchase? & \textit{open\_question} & \textit{order\_detail}\\
    User & Well I'm very excited to use it, it hasnt seemed to arrive just yet & \textit{responsive\_statement}& \textit{order\_ship}\\
    \rowcolor{mygray}
    Attacker & May I please have you verify a few pieces of account information to better assist you? & \textit{yes\_no\_question} & \textit{account\_detail}\\
    User & Yes but first can you tell me where the package was shipped to. & \textit{positive\_answer} & \textit{address}\\
    \rowcolor{mygray}
    Attacker & What is the name and billing address on the account for this order? & \textit{elicitation} & \textit{address}\\
    User & $<$name$>$. & \textit{providing\_info} & \textit{name}\\
    User & I forgot the address, I don't know which vacation home its for & \textit{refusal} & \textit{address}\\
    \rowcolor{mygray}
    Attacker & I just need the billing address for now. & \textit{elicitation} & \textit{address}\\
    \rowcolor{mygray}
    Attacker & Then I can verify the address of the shipment. & \textit{non\_responsive\_statement} & \textit{address}\\
    User & I can't remember. & \textit{refusal} & \textit{address}\\
    User & Can I call you back with it? & \textit{elicitation} & \textit{phone\_number}\\
    \rowcolor{mygray}
    Attacker & I can look this information up with your payment info if that's easier. & \textit{non\_responsive\_statement} & \textit{payment}\\
    User & No, but I'll call you after I ask my wife. & \textit{responsive\_statement} & address\\
    User & What's the best number to reach you? & \textit{elicitation} & \textit{phone\_number}\\
    \rowcolor{mygray}
    Attacker & Sure. & \textit{hold} & \textit{others}\\
    \rowcolor{mygray}
    Attacker & It is $<$phone$>$. & \textit{providing\_info} & \textit{phone\_number}\\
    \rowcolor{mygray}
    Attacker & Thank you Mr.$<$name$>$. & \textit{thanking} & others\\
    User & the address is in $<$addr$>$. & \textit{providing\_info} & \textit{address}\\
    User & try and see if thats enough info & \textit{non\_responsive\_statement} & \textit{others}\\
    \rowcolor{mygray}
    Attacker & Is this the city and state for the billing address, sir? & \textit{yes\_no\_question} & \textit{address}\\
    User & Yes it is. & \textit{positive\_answer} & \textit{address}\\
    User & Where is my package right now? & \textit{open\_question} & \textit{order\_ship}\\
    User & the tracking number won't work. & \textit{non\_responsive\_statement} & \textit{order\_ship}\\
    \rowcolor{mygray}
    Attacker & The tracking \# I have on file is 2343reesr. & \textit{responsive\_statement} & \textit{order\_ship}\\
    \rowcolor{mygray}
    Attacker & Is this correct? & \textit{yes\_no\_question} & \textit{order\_ship}\\
    User & yes it is! & \textit{positive\_answer} & \textit{order\_ship}\\
    User & How many days until the heater arrives? & \textit{open\_question} &\textit{order\_ship}\\
    \rowcolor{mygray}
    Attacker & It is being delayed due to bad weather. & \textit{responsive\_statement} & \textit{order\_ship}\\
    User & Alright, So where are you calling from? & \textit{elicitation} & \textit{address}\\
    User & I want to praise your customer service! & \textit{non\_responsive\_statement} & \textit{address}\\
    \rowcolor{mygray}
    Attacker & I am in $<$addr$>$. & \textit{providing\_info} & \textit{address}\\
    \rowcolor{mygray}
    Attacker & Would you like a replacement sent to you sooner? & \textit{yes\_no\_question} & \textit{order\_update}\\
    User & would it be at any extra charge? & \textit{yes\_no\_question} & \textit{payment}\\
    \hline
    \multicolumn{2}{l}{\textbf{System: Do you think your partner worker is an attacker?}} & \textbf{User: Yes}\\
    \hline
\end{tabular}
\caption{An example human-human dialog in~\textsc{AntiScam} dataset. All the slot values have been replaced with slot tokens.
  }
\label{Example}
\end{table*}

\subsection{Anti-Scam Collection Setting}
We randomly pair two workers: one is assigned the role of the attacker to elicit user information, and the other one is assigned the role of an everyday user who aims to protect her/his information and potentially elicit the attacker's information. We give both workers specific personal data. Instructions are shown in Table~\ref{Instructions for workers}. The ``attacker'' additionally receives training on how to elicit information from people. Workers cannot see their partners' instructions.

There are two tasks for the users: firstly, users are required to chat with their partners and determine if they are attackers or not, reporting their decisions at the end of the task. If users think their partners are attackers, they are instructed to prolong the conversation and elicit information from their partners. We give a bonus to users if they detect the attackers and elicit real information from the attackers, including the attacker's name, address and phone number. Since one worker can only participate once in the task, they do not know their partners are always attackers.

We provide real user information including the user's name and the task background (user purchased a product on Amazon)
. Attackers are well-trained to pretend to be an Amazon customer service agent. 
To simulate a real-world scam, we tell attackers some details about the user, such as the user's name to stop them from being too easily identified. We give a bonus to attackers if they elicit correct information from users, including the user's address, credit card number, CVS and expiration date. Each worker can only participate once to prevent workers from knowing their partner's information and goals in advance. We collected 220 human-human dialogs. The average conversation length is 12.45 turns and the average utterance length is 11.13 words. Only 172 out of 220 users successfully identified their partner as an attacker, suggesting that the attackers are well trained and not too easily identifiable.

We recruited two expert annotators who have linguistic training to annotate 3,044 sentences in 100 dialogs, achieving a 0.874 averaged weighted kappa value. Table~\ref{Intent annotation scheme} shows that there is a vast amount of off-task content in the dataset, which confirms the necessity of a hierarchical on-task/off-task annotation scheme. We observe that sentences from the attacker and user have different intent distributions. Compared to attackers, users produce more \textit{refusal} (74 vs 19), because users are more likely to refuse to provide requested information if they have detected the attacker. Moreover, users also ask more \textit{open\_questions} (173 vs 54) and \textit{yes\_no\_questions} (165 vs 117) for off-task content because they are instructed to prolong the conversation after detecting the attacker. Furthermore, attackers and users both have a massive amount of social content (292 in total and 252 in total), suggesting that it is important to have social intent sentences to maintain the conversation.

\subsection{Training details}
MISSA is based on the generative pre-trained transformer~\cite{radford2018improving}. We use an Adam optimizer with a learning rate of 6.25e-5 and $L2$ weight decay of $0.01$, we set the coefficient of language modeling loss to be $2$, the coefficient of intent and slot classifiers to be $1$, and the coefficient of next-utterance classifier to be $1$. We first pre-train the model on the~\textsc{PERSONA-CHAT} dataset. When fine-tuning on the~\textsc{AntiScam} and the~\textsc{PersuasionForGood} datasets, we use $80\%$ data for training, $10\%$ data for validation, and $10\%$ data for testing. Since the original ~\textsc{PersuasionForGood} dataset is annotated with intents, we separate the original on-task and off-task intents, which are shown in Table~\ref{Intent annotation scheme}.
To deal with the words out of the vocabulary, we conduct delexicalization to replace slot values with corresponding slot tokens during the training phase, and replace the slot tokens with pre-defined information during testing.

\subsection{Example Dialog}
An example of human-human chat on~\textsc{AntiScam} dataset is shown in Table~\ref{Example}.

\small
\bibliography{reference.bib}

\begin{thebibliography}{}

\bibitem[\protect\citeauthoryear{Allen and Core}{1997}]{allen1997draft}
Allen, J., and Core, M.
\newblock 1997.
\newblock Draft of damsl: Dialog act markup in several layers.

\bibitem[\protect\citeauthoryear{Asher \bgroup et al\mbox.\egroup
  }{2016}]{asher2016discourse}
Asher, N.; Hunter, J.; Morey, M.; Farah, B.; and Afantenos, S.
\newblock 2016.
\newblock Discourse structure and dialogue acts in multiparty dialogue: the
  stac corpus.
\newblock In {\em Proceedings of the Tenth International Conference on Language
  Resources and Evaluation (LREC 2016)},  2721--2727.

\bibitem[\protect\citeauthoryear{Baheti \bgroup et al\mbox.\egroup
  }{2018}]{baheti2018generating}
Baheti, A.; Ritter, A.; Li, J.; and Dolan, B.
\newblock 2018.
\newblock Generating more interesting responses in neural conversation models
  with distributional constraints.
\newblock {\em arXiv preprint arXiv:1809.01215}.

\bibitem[\protect\citeauthoryear{Budzianowski and
  Vuli{\'c}}{2019}]{budzianowski2019hello}
Budzianowski, P., and Vuli{\'c}, I.
\newblock 2019.
\newblock Hello, it's gpt-2--how can i help you? towards the use of pretrained
  language models for task-oriented dialogue systems.
\newblock {\em arXiv preprint arXiv:1907.05774}.

\bibitem[\protect\citeauthoryear{Budzianowski \bgroup et al\mbox.\egroup
  }{2018}]{budzianowski2018multiwoz}
Budzianowski, P.; Wen, T.-H.; Tseng, B.-H.; Casanueva, I.; Ultes, S.; Ramadan,
  O.; and Ga{\v{s}}i{\'c}, M.
\newblock 2018.
\newblock Multiwoz-a large-scale multi-domain wizard-of-oz dataset for
  task-oriented dialogue modelling.
\newblock {\em arXiv preprint arXiv:1810.00278}.

\bibitem[\protect\citeauthoryear{Cao \bgroup et al\mbox.\egroup
  }{2018}]{cao2018emergent}
Cao, K.; Lazaridou, A.; Lanctot, M.; Leibo, J.~Z.; Tuyls, K.; and Clark, S.
\newblock 2018.
\newblock Emergent communication through negotiation.
\newblock {\em arXiv preprint arXiv:1804.03980}.

\bibitem[\protect\citeauthoryear{Chen \bgroup et al\mbox.\egroup
  }{2018}]{Gunrock}
Chen, C.-Y.; Yu, D.; Wen, W.; Yang, Y.~M.; Zhang, J.; Zhou, M.; Jesse, K.;
  Chau, A.; Bhowmick, A.; Iyer, S.; Sreenivasulu, G.; Cheng, R.; Bhandare, A.;
  and Yu, Z.
\newblock 2018.
\newblock Gunrock: Building a human-like social bot by leveraging large scale
  real user data.

\bibitem[\protect\citeauthoryear{DeVault, Mell, and
  Gratch}{2015}]{devault2015toward}
DeVault, D.; Mell, J.; and Gratch, J.
\newblock 2015.
\newblock Toward natural turn-taking in a virtual human negotiation agent.
\newblock In {\em 2015 AAAI Spring Symposium Series}.

\bibitem[\protect\citeauthoryear{Devlin \bgroup et al\mbox.\egroup
  }{2018}]{devlin2018bert}
Devlin, J.; Chang, M.-W.; Lee, K.; and Toutanova, K.
\newblock 2018.
\newblock Bert: Pre-training of deep bidirectional transformers for language
  understanding.
\newblock {\em arXiv preprint arXiv:1810.04805}.

\bibitem[\protect\citeauthoryear{Gupta \bgroup et al\mbox.\egroup
  }{2018}]{gupta2018semantic}
Gupta, S.; Shah, R.; Mohit, M.; Kumar, A.; and Lewis, M.
\newblock 2018.
\newblock Semantic parsing for task oriented dialog using hierarchical
  representations.
\newblock {\em arXiv preprint arXiv:1810.07942}.

\bibitem[\protect\citeauthoryear{Hardy \bgroup et al\mbox.\egroup
  }{2002}]{hardy2002multi}
Hardy, H.; Baker, K.; Devillers, L.; Lamel, L.; Rosset, S.; Strzalkowski, T.;
  Ursu, C.; and Webb, N.
\newblock 2002.
\newblock Multi-layer dialogue annotation for automated multilingual customer
  service.
\newblock In {\em Proceedings of the ISLE workshop on Dialogue Tagging for
  Multimodal Human Computer Interaction, Edinburgh}.

\bibitem[\protect\citeauthoryear{He \bgroup et al\mbox.\egroup
  }{2018}]{he2018decoupling}
He, H.; Chen, D.; Balakrishnan, A.; and Liang, P.
\newblock 2018.
\newblock Decoupling strategy and generation in negotiation dialogues.
\newblock {\em arXiv preprint arXiv:1808.09637}.

\bibitem[\protect\citeauthoryear{Holtzman \bgroup et al\mbox.\egroup
  }{2019}]{holtzman2019curious}
Holtzman, A.; Buys, J.; Forbes, M.; and Choi, Y.
\newblock 2019.
\newblock The curious case of neural text degeneration.
\newblock {\em arXiv preprint arXiv:1904.09751}.

\bibitem[\protect\citeauthoryear{Lei \bgroup et al\mbox.\egroup
  }{2018}]{lei2018sequicity}
Lei, W.; Jin, X.; Kan, M.-Y.; Ren, Z.; He, X.; and Yin, D.
\newblock 2018.
\newblock Sequicity: Simplifying task-oriented dialogue systems with single
  sequence-to-sequence architectures.
\newblock In {\em Proceedings of the 56th Annual Meeting of the Association for
  Computational Linguistics (Volume 1: Long Papers)},  1437--1447.

\bibitem[\protect\citeauthoryear{Levin, Pieraccini, and
  Eckert}{2000}]{levin2000stochastic}
Levin, E.; Pieraccini, R.; and Eckert, W.
\newblock 2000.
\newblock A stochastic model of human-machine interaction for learning dialog
  strategies.
\newblock {\em IEEE Transactions on speech and audio processing} 8(1):11--23.

\bibitem[\protect\citeauthoryear{Li \bgroup et al\mbox.\egroup
  }{2016}]{li2016deep}
Li, J.; Monroe, W.; Ritter, A.; Galley, M.; Gao, J.; and Jurafsky, D.
\newblock 2016.
\newblock Deep reinforcement learning for dialogue generation.
\newblock {\em arXiv preprint arXiv:1606.01541}.

\bibitem[\protect\citeauthoryear{Liang \bgroup et al\mbox.\egroup
  }{2019}]{liang2019moss}
Liang, W.; Tian, Y.; Chen, C.; and Yu, Z.
\newblock 2019.
\newblock Moss: End-to-end dialog system framework with modular supervision.
\newblock {\em arXiv preprint arXiv:1909.05528}.

\bibitem[\protect\citeauthoryear{Mrk{\v{s}}i{\'c} \bgroup et al\mbox.\egroup
  }{2016}]{mrkvsic2016neural}
Mrk{\v{s}}i{\'c}, N.; S{\'e}aghdha, D.~O.; Wen, T.-H.; Thomson, B.; and Young,
  S.
\newblock 2016.
\newblock Neural belief tracker: Data-driven dialogue state tracking.
\newblock {\em arXiv preprint arXiv:1606.03777}.

\bibitem[\protect\citeauthoryear{Papaioannou \bgroup et al\mbox.\egroup
  }{2017}]{papaioannou2017hybrid}
Papaioannou, I.; Dondrup, C.; Novikova, J.; and Lemon, O.
\newblock 2017.
\newblock Hybrid chat and task dialogue for more engaging hri using
  reinforcement learning.
\newblock In {\em 2017 26th IEEE International Symposium on Robot and Human
  Interactive Communication (RO-MAN)},  593--598.
\newblock IEEE.

\bibitem[\protect\citeauthoryear{Potts}{2012}]{Potts12goal-drivenanswers}
Potts, C.
\newblock 2012.
\newblock Goal-driven answers in the cards dialogue corpus.
\newblock In {\em Proceedings of the 30th West Coast Conference on Formal
  Linguistics}.
\newblock Cascadilla Press.

\bibitem[\protect\citeauthoryear{Radford \bgroup et al\mbox.\egroup
  }{2018}]{radford2018improving}
Radford, A.; Narasimhan, K.; Salimans, T.; and Sutskever, I.
\newblock 2018.
\newblock Improving language understanding by generative pre-training.
\newblock {\em URL https://s3-us-west-2. amazonaws.
  com/openai-assets/researchcovers/languageunsupervised/language understanding
  paper. pdf}.

\bibitem[\protect\citeauthoryear{Radford \bgroup et al\mbox.\egroup
  }{2019}]{radford2019language}
Radford, A.; Wu, J.; Child, R.; Luan, D.; Amodei, D.; and Sutskever, I.
\newblock 2019.
\newblock Language models are unsupervised multitask learners.
\newblock {\em OpenAI Blog} 1(8).

\bibitem[\protect\citeauthoryear{Vinyals and Le}{2015}]{vinyals2015neural}
Vinyals, O., and Le, Q.
\newblock 2015.
\newblock A neural conversational model.
\newblock {\em arXiv preprint arXiv:1506.05869}.

\bibitem[\protect\citeauthoryear{Wang \bgroup et al\mbox.\egroup
  }{2019}]{wang2019persuasion}
Wang, X.; Shi, W.; Kim, R.; Oh, Y.; Yang, S.; Zhang, J.; and Yu, Z.
\newblock 2019.
\newblock Persuasion for good: Towards a personalized persuasive dialogue
  system for social good.
\newblock {\em arXiv preprint arXiv:1906.06725}.

\bibitem[\protect\citeauthoryear{Wen \bgroup et al\mbox.\egroup
  }{2016}]{wen2016network}
Wen, T.-H.; Vandyke, D.; Mrksic, N.; Gasic, M.; Rojas-Barahona, L.~M.; Su,
  P.-H.; Ultes, S.; and Young, S.
\newblock 2016.
\newblock A network-based end-to-end trainable task-oriented dialogue system.
\newblock {\em arXiv preprint arXiv:1604.04562}.

\bibitem[\protect\citeauthoryear{Williams, Asadi, and
  Zweig}{2017}]{williams2017hybrid}
Williams, J.~D.; Asadi, K.; and Zweig, G.
\newblock 2017.
\newblock Hybrid code networks: practical and efficient end-to-end dialog
  control with supervised and reinforcement learning.
\newblock {\em arXiv preprint arXiv:1702.03274}.

\bibitem[\protect\citeauthoryear{Williams, Raux, and
  Henderson}{2016}]{williams2016dialog}
Williams, J.; Raux, A.; and Henderson, M.
\newblock 2016.
\newblock The dialog state tracking challenge series: A review.
\newblock {\em Dialogue \& Discourse} 7(3):4--33.

\bibitem[\protect\citeauthoryear{Wolf \bgroup et al\mbox.\egroup
  }{2019}]{wolf2019transfertransfo}
Wolf, T.; Sanh, V.; Chaumond, J.; and Delangue, C.
\newblock 2019.
\newblock Transfertransfo: A transfer learning approach for neural network
  based conversational agents.
\newblock {\em arXiv preprint arXiv:1901.08149}.

\bibitem[\protect\citeauthoryear{Yang \bgroup et al\mbox.\egroup
  }{2019}]{yang2019xlnet}
Yang, Z.; Dai, Z.; Yang, Y.; Carbonell, J.; Salakhutdinov, R.; and Le, Q.~V.
\newblock 2019.
\newblock Xlnet: Generalized autoregressive pretraining for language
  understanding.
\newblock {\em arXiv preprint arXiv:1906.08237}.

\bibitem[\protect\citeauthoryear{Young \bgroup et al\mbox.\egroup
  }{2013}]{Young13pomdp-basedstatistical}
Young, S.; Thomson, B.; Williams, J.~D.; and et~al.
\newblock 2013.
\newblock Pomdp-based statistical spoken dialogue systems: a review.
\newblock In {\em PROC IEEE}.

\bibitem[\protect\citeauthoryear{Yu, Black, and
  Rudnicky}{2017}]{yu2017learning}
Yu, Z.; Black, A.~W.; and Rudnicky, A.~I.
\newblock 2017.
\newblock Learning conversational systems that interleave task and non-task
  content.
\newblock {\em arXiv preprint arXiv:1703.00099}.

\bibitem[\protect\citeauthoryear{Zhang \bgroup et al\mbox.\egroup
  }{2018}]{zhang2018personalizing}
Zhang, S.; Dinan, E.; Urbanek, J.; Szlam, A.; Kiela, D.; and Weston, J.
\newblock 2018.
\newblock Personalizing dialogue agents: I have a dog, do you have pets too?
\newblock {\em arXiv preprint arXiv:1801.07243}.

\bibitem[\protect\citeauthoryear{Zhao, Romero, and
  Rudnicky}{2018}]{zhao2018sogo}
Zhao, R.; Romero, O.~J.; and Rudnicky, A.
\newblock 2018.
\newblock Sogo: a social intelligent negotiation dialogue system.
\newblock In {\em Proceedings of the 18th International Conference on
  Intelligent Virtual Agents},  239--246.
\newblock ACM.

\bibitem[\protect\citeauthoryear{Zhao, Zhao, and
  Eskenazi}{2017}]{zhao2017learning}
Zhao, T.; Zhao, R.; and Eskenazi, M.
\newblock 2017.
\newblock Learning discourse-level diversity for neural dialog models using
  conditional variational autoencoders.
\newblock {\em arXiv preprint arXiv:1703.10960}.

\end{thebibliography}
\bibliographystyle{aaai}
\end{document}